%% file: main.tex
\pdfoutput=1
\documentclass[11pt]{article}
\input{macro}

\usepackage{emnlp2022}

\usepackage{times}
\usepackage{latexsym}

\usepackage[T1]{fontenc}
\usepackage[utf8]{inputenc}

\usepackage{microtype}

\title{Context-Situated Pun Generation}

\author{Jiao Sun$^1$\thanks{\xspace\xspace Work done during Jiao's internship at Amazon.} \, Anjali Narayan-Chen$^{2}$ \, Shereen Oraby$^{2}$ \, Shuyang Gao$^{2}$\thanks{\xspace\xspace Work done while Shuyang was at Amazon.}  \,\\  \textbf{Tagyoung Chung}$^{2}$ \, \textbf{Jing Huang}$^{2}$ \, \textbf{Yang Liu}$^{2}$ \, \textbf{Nanyun Peng}$^{2,3}$ \\
$^1$University of Southern California \\
$^2$Amazon Alexa AI\\
$^3$University of California, Los Angeles \\
\texttt{jiaosun@usc.edu} \\
\texttt{\{naraanja,orabys,shuyag,tagyoung,jhuangz,yangliud\}@amazon.com }\\
\texttt{violetpeng@cs.ucla.edu}
}

\begin{document}
\maketitle
\begin{abstract}
\input{content/0-abstract}
\end{abstract}

\input{content/1-intro}
\input{content/3-formulation}

\input{content/4-model}
\input{content/5-experiment}
\input{content/related}

\input{content/99-conclusion}

\input{content/limitations}

% Entries for the entire Anthology, followed by custom entries
\bibliography{anthology, cpun}
\bibliographystyle{acl_natbib}
\clearpage
\appendix
\input{appendix/modeling_details}

\input{appendix/retrieval_examples}
\input{appendix/annotation-guideline}

\end{document}

%% file: macro.tex
\usepackage{color}
\definecolor{wildstrawberry}{rgb}{1.0, 0.26, 0.64}
\newcommand{\datasetname}{CUP\xspace}

\usepackage{times}
\usepackage{subcaption}
\usepackage{latexsym}
\usepackage{xspace}
\usepackage{booktabs}
\usepackage{multirow}
\usepackage{graphicx}
\usepackage{tikz}
\usetikzlibrary{plotmarks}
\usepackage{fontawesome5}
\usepackage{quoting}
\usepackage{multirow, makecell}

\usepackage{algorithm}
\usepackage{algorithmic}
\usepackage{soul}
\usepackage{enumitem}
\usepackage{todonotes}
\usepackage{fontawesome5}

\newcommand{\RNum}[1]{\uppercase\expandafter{\romannumeral #1\relax}}

%% file: content/0-abstract.tex
Previous work on pun generation commonly begins with a given pun word (a pair of homophones for heterographic pun generation and a polyseme for homographic pun generation) and seeks to generate an appropriate pun. While this may enable efficient pun generation, we believe that a pun is most entertaining if it fits appropriately within a given context, e.g., a given situation or dialogue. In this work, we propose a new task, \textit{context-situated pun generation}, where a specific context represented by a set of keywords is provided, and the task is to first identify suitable pun words that are appropriate for the context, then generate puns based on the context keywords and the identified pun words. 
We collect {\faMugHot} {\datasetname} (\textbf{C}ontext-sit\textbf{U}ated \textbf{P}un), containing 4.5k tuples of context words and pun pairs. Based on the new data and setup, we propose a pipeline system for context-situated pun generation, including a pun word retrieval module that identifies suitable pun words for a given context, and a generation module that generates puns from context keywords and pun words. 
Human evaluation shows that 69\% of our top retrieved pun words can be used to generate context-situated puns, and our generation module yields successful puns 31\% of the time given a plausible tuple of context words and pun pair, almost tripling the yield of a state-of-the-art pun generation model. With an end-to-end evaluation, our pipeline system with the top-1 retrieved pun pair for a given context can generate successful puns 40\% of the time, better than all other modeling variations but 32\% lower than the human success rate. This highlights the difficulty of the task, and encourages more research in this direction.

% First, we propose a new task, context situated pun generation, in the field of pun generation. 
% The task is built on our belief that a pun will only be funny given a correct context. 
% Compared to previous pun generation works that start from a given pun word (a pair of homophones for heterographic pun generation and a polyseme for homographic pun generation), our work first asks the question of ``under a specific context, what pun words would be appropriate to make a spontaneous pun?'' Second, we propose a pipeline system for context-situated pun generation. The system contains a pun word retrieval module that can identify suitable pun words for a given context and a pun generation module. Our human evaluation shows that 69\% of our retrieved pun words can be used to generate puns. Our pun generation module unifies the heterographic pun generation and the homographic pun generation for the first time.  Third, we collect a dataset that contain 4,551 context words and pun word pairs, with the label of if they are compatible to make a pun. At the same time, we will release a corpus of 2,692 high-quality human-written puns for the compatible context words and pun words. We benchmark the model performance on our dataset to establish a strong baseline. As the pilot work introducing the \textbf{context-situated} concept into the pun generation community, we hope that our work can spark context-situated innovation on other domains of creative generation. 

%% file: content/1-intro.tex
\section{Introduction}
\begin{figure}[t]
    \centering
    \includegraphics[width=\linewidth]{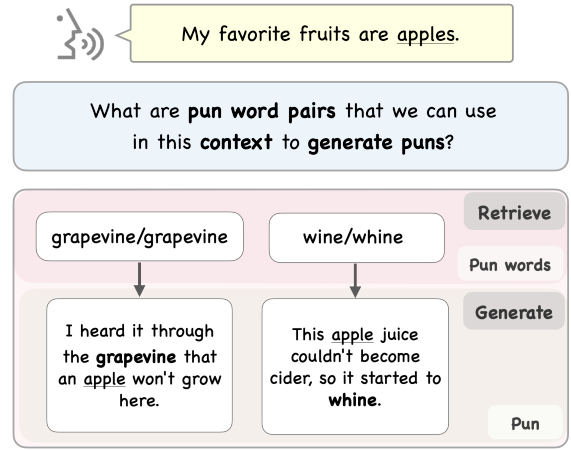}
    \caption{\emph{Context-situated pun generation} aims to find relevant pun words to generate puns within a given context. We propose a unified framework to generate both homographic and heterographic puns; examples shown here are human-written puns from our corpus. 
    % We use \underline{underline} to highlight the context and \textbf{bold} to highlight pun words.
     }
    \label{fig:teaser}
\end{figure}

\begin{table*}[]
\small
\renewcommand{\tabcolsep}{1.4mm}
\resizebox{\textwidth}{!}{
\begin{tabular}{@{}clllll@{}}
\toprule
\textbf{Type} &
  \textbf{Pun} &
  \textbf{$p_w$/$a_w$} &
  \textbf{Context $C$} &
  \textbf{$S_{p_w}$} &
  \textbf{$S_{a_w}$} \\ \midrule
\multirow{4}{*}{het.} &
  \begin{tabular}[c]{@{}l@{}}Two construction workers \\ had a staring contest.\end{tabular} &
  \begin{tabular}[c]{@{}l@{}}stair/ \\ stare\end{tabular} &
  \begin{tabular}[c]{@{}l@{}}construction \\ workers\end{tabular} &
  \begin{tabular}[c]{@{}l@{}}support consisting of a place \\ to rest the foot while ascending \\ or descending a stairway\end{tabular} &
  look at with fixed eyes \\  \cmidrule{2-6}
  & \begin{tabular}[c]{@{}l@{}}``I've stuck a pin through my \\ nose'', said Tom punctually.\end{tabular} &
  \begin{tabular}[c]{@{}l@{}}punctually/ \\puncture\end{tabular} &
  pin, nose &
  at the expected or proper time &
  \begin{tabular}[c]{@{}l@{}}a small hole made\\ by a sharp object\end{tabular} \\ \midrule
\multirow{4}{*}{hom.} &
  \begin{tabular}[c]{@{}l@{}}A new type of broom came \\out, it is sweeping the country.\end{tabular} &
  \begin{tabular}[c]{@{}l@{}}sweep/ \\ sweep\end{tabular} &
  \begin{tabular}[c]{@{}l@{}}broom, \\ nation\end{tabular} &
  \begin{tabular}[c]{@{}l@{}} sweep with a broom or as if \\ with a broom \end{tabular} &
  \begin{tabular}[c]{@{}l@{}}win an overwhelming\\ victory in or on\end{tabular} \\  \cmidrule{2-6} 
 &
  \begin{tabular}[c]{@{}l@{}}If you sight a whale, it could \\ be a fluke.\end{tabular} &
  \begin{tabular}[c]{@{}l@{}}fluke/\\fluke\end{tabular} &
  whale &
  a stroke of luck &
  \begin{tabular}[c]{@{}l@{}}either of the two lobes \\ of the tail of a cetacean\end{tabular} \\ \bottomrule
\end{tabular}
}
\caption{Two examples each of heterographic puns and homographic puns in the SemEval 2017 Task 7 dataset. 
We construct context $C$ by extracting keywords from the pun and excluding the pun word $p_w$. Word sense information $S_{p_w}$ and $S_{a_w}$ are retrieved from WordNet from SemEval annotated senses.}
% After extracting the keywords from the pun, we exclude the pun word from the keywords as a proxy for the context information. 
% We get the sense information from WordNet after looking up the sense key in the SemEval annotation.}
\label{tab:examples}
\end{table*}

Pun generation is a challenging creative generation task that has attracted some recent attention in the research community~\cite{he-etal-2019-pun,yu-etal-2018-neural,yu-etal-2020-homophonic,mittal2022ambipun,horri2011linguistic}. 
As one of the most important ways to communicate humor~\cite{abbas2016pun}, puns can help relieve anxiety, avoid painful feelings and facilitate learning~\cite{buxman2008humor}. 
At the same time, spontaneity is the twin concept of creativity~\cite{moreno1955theory}, which means the context matters greatly for making an appropriate and funny pun. 

Existing work on pun generation mainly focuses on generating puns given a pair of pun-alternative words or senses (we call it a pun pair). Specifically, in heterographic pun generation, systems generate puns using a pair of homophones involving a pun word and an alternative word~\cite{he-etal-2019-pun, yu-etal-2020-homophonic, mittal2022ambipun}.
% The systems are supposed to generate puns containing the pun word and also indicate the alternative word.
Alternatively, in homographic pun generation, systems generate puns that must support both given senses of a single polysemous word~\cite{yu-etal-2018-neural, luo-etal-2019-pun, tian2022unified}. 
% The generated puns need to support both given senses of the polysemous pun word.
Despite the great progress that has been made under such experimental settings, real-world applications for pun generation (e.g., in dialogue systems or creative slogan generation) rarely have these pun pairs provided. 
Instead, puns need to be generated given a more naturally-occurring conversational or creative context, requiring the identification of a pun pair that is relevant and appropriate for that context. For example, given a conversation turn \textit{``How was the magic show?''}, a context-situated pun response might be, \textit{``The magician got so mad he pulled his hare out.''}
% Instead, there is usually a conversational or general context that exists, and a pun-alternative word/sense pair should be identified under that context to compose puns.

Motivated by real-world applications and the theory that the funniness of a pun heavily relies on the context, we formally define and introduce a new setting for pun generation, which we call \textit{context-situated pun generation}: given a context represented by a set of keywords, the task is to generate puns that fit the given context (Figure~\ref{fig:teaser}). %Furthermore, we build a pipeline system to generate context-situated puns to establish a strong baseline for this task. The system contains a \emph{retrieval module} that finds a set of pun words that are suitable for making a pun given the current context, and a \emph{generation module} that generates puns from the context words and the retrieved pun word.} 
Our contributions are as follows:
\begin{itemize}
% \vspace{-0.1in}
    % Jing: no need to repeat it again here.
    \item We introduce a new setting of \textit{context situated pun generation}. %, where, given a context represented as a set of keywords, the task is to infer appropriate pun-alternative word or sense pairs for the given context and generate puns subsequently.
    \item To facilitate research in this direction, we collect a large-scale corpus called {\faMugHot} {\datasetname} (\textbf{C}ontext-sit\textbf{U}ated \textbf{P}un), which contains 4,551 tuples of context keywords and an associated pun pair, each labelled with whether they are compatible for composing a pun. If a tuple is compatible, we additionally collect a human-written pun that incorporates both the context keywords and the pun word.\footnote{Resources will be available at: \\ \url{https://github.com/amazon-research/context-situated-pun-generation}}
    \item We build a pipeline system with a {\it retrieval module} to predict proper pun words given the current context, and a {\it generation} module to incorporate both the context keywords and the pun word to generate puns. Our system serves as a strong baseline for \textit{context situated pun generation}.  %Human evaluation indicates that humans can come up with a pun given the context words and predicted pun words for 69\% time. %For the pun generation component, we propose a unified framework for generating both homographic and homophonic puns.
\end{itemize}

%For example, the classic pun ``The magician got so mad he pulled his hare out.'' would be funnier if it happens as part of the conversation in a magic show. Motivated by the intuition that the funniness of a pun heavily relies on the context, this work asks the questions of ``under a specific context, what pun words would be appropriate to make a pun spontaneously?'' for the first time in the pun generation community. We introduce this setting where generated puns should situate in a given context as \textbf{context-situated pun generation}.

%The most state-of-the pun generation works on pun generation mainly focus on either homophonic pun generation~\cite{he-etal-2019-pun, yu-etal-2020-homophonic} or homographic pun generation~\cite{mittal2022ambipun, luo-etal-2019-pun}. In all of these works, the pun words are pre-given as a starting point. For homophonic puns, the pun word will be one of a homophonic pair, while for homographic puns, the pun word will be a polyseme which has two or more meanings. Despite of the great progress researchers have been making so far, we see the current setting as an experimental setting which is less practical. In a real world application for pun generation, we do not know the pun words beforehand, and all we know is the context instead. Then, we need to infer suitable pun words for the given context. After that, we will generate a pun using the inferred pun words and context words. 

%% file: content/3-formulation.tex
\section{Task Formulation}
\paragraph{Preliminaries.} Ambiguity is the key to pun generation~\cite{ritchie-2005-computational}. First, we define the term \emph{pun pair} in our work. For heterographic pun generation, there exists a pair of homophones, which we call \emph{pun word} ($p_w$) and \emph{alternative word} ($a_w$). While only $p_w$ appears in the pun, both the meaning of $p_w$ and $a_w$ are supported in the pun sentence. Therefore, the input of heterographic pun generation can be written as \textrm{(\textbf{$p_w$}, $S_{p_w}$, \textbf{$a_w$}, $S_{a_w}$)},
% \begin{equation}
%     \textrm{(\textbf{$p_w$}, $S_{p_w}$, \textbf{$a_w$}, $S_{a_w}$)}.
%     \label{eq:def}
% \end{equation}
where $S_{p_w}$ and $S_{a_w}$ are the senses of the pun word and alternative word, respectively. We refer to these as \textbf{\emph{pun pairs}}, and use the shorthand $(p_w, a_w)$ for simplicity. For homographic pun generation, the pun word is a polyseme that has two meanings; here, we can use the same representation, where $p_w=a_w$ for homographic puns.

\paragraph{Formulation.} Given the unified representation for heterographic and homographic puns, we define the task of context-situated pun generation as: {\it Given a context $C$, which can be a sentence or a list of keywords, find a pun pair ($p_w$, $S_{p_w}$, $a_w$, $S_{a_w}$) that is suitable to generate a pun, then generate a pun using the chosen pun pair situated in the given context}. In this work, we assume we are given a fixed set of pun pair candidates $(P_w, A_w)$ from which ($p_w$, $a_w$) are retrieved. The unified format between heterographic and heterographic puns makes it possible for us to propose a unified framework for pun generation.
% later in this paper.  %% ANJALI

\begin{table}[]
\small\centering
\resizebox{\columnwidth}{!}{
\begin{tabular}{@{}lll@{}}
\toprule
\textbf{$p_w$ / $a_w$} & $L$ & \multicolumn{1}{c}{\textbf{Context-Situated Pun} for \textit{hunts, deer}} \\ \midrule
\begin{tabular}[c]{@{}l@{}}hedges/ \\ edges\end{tabular} & 1 & \begin{tabular}[c]{@{}l@{}}Why is the hunter so good at hunting deer? \\Because he hunts life on the hedges\end{tabular} \\ \midrule
husky/husk & 0 & - \\ \midrule
\begin{tabular}[c]{@{}l@{}}catch/ \\ catch\end{tabular} & 1 & \begin{tabular}[c]{@{}l@{}}He hunts deer but the catch is that they \\ rarely show up.\end{tabular} \\ \midrule
\begin{tabular}[c]{@{}l@{}}pine/ \\ pine\end{tabular} & 1 & \begin{tabular}[c]{@{}l@{}}Hunting deer in the forest always makes \\ him pine for the loss.\end{tabular} \\ \midrule
\begin{tabular}[c]{@{}l@{}}boar/ \\ bore\end{tabular} & 1 & \begin{tabular}[c]{@{}l@{}}He is so mundane about hunting deer, \\ but it is hardly a boar.\end{tabular} \\ \midrule
\begin{tabular}[c]{@{}l@{}}jerky/ \\ jerky\end{tabular} & 1 & \begin{tabular}[c]{@{}l@{}}What do you call an erratic deer that is \\ being hunted? Jerky\end{tabular} \\ \bottomrule
\end{tabular}}
\caption{Example annotations from the {\faMugHot} {\datasetname} dataset. Labels $L$ indicate whether the annotator was able to write a pun given the context and pun pair.}

\label{tab:annotation}
\end{table}

\section{{\faMugHot} {\datasetname} Dataset}
\paragraph{Motivation.} The largest and most commonly-used dataset in the pun generation community is the SemEval 2017 Task 7 dataset~\cite{miller-etal-2017-semeval}.\footnote{\url{https://alt.qcri.org/semeval2017/task7/}. The data is released under CC BY-NC 4.0 license (\url{https://creativecommons.org/licenses/by-nc/4.0/legalcode}).} Under our setting of context-situated pun generation, we can utilize keywords from the puns themselves as context.  However, the majority of pun pairs only occur once in the the SemEval dataset, while one given context could have been compatible with many other pun pairs. For example, given the context \emph{\underline{beauty school, class}}, the original pun in the SemEval dataset uses the homographic pun pair \emph{(makeup, makeup)} and says: \emph{``If you miss a \underline{class} at \underline{beauty school} you'll need a \emph{makeup} session.''} At the same time, a creative human can use the heterographic pun pair \emph{(dyed, die)} to instead generate \emph{``I inhaled so much ash from the eye shadow palette at the \underline{beauty school class} -- I might have \emph{dyed} a little inside.''} 
Because of the limitation of the SemEval dataset, we need a dataset that has a diverse set of pun pairs combined with given contexts. Furthermore, the dataset should be annotated to indicate whether the context words and pun pair combination is suitable to make context-situated puns. \looseness=-1 %a pun pair can be used for making a pun that is situated in a given context.

\paragraph{Data Preparation.} We sample puns that contain both sense annotations and pun word annotations from SemEval Task 7.
% ~\cite{miller-etal-2017-semeval}. 
We show two examples of heterographic puns and homographic puns and their annotations from the SemEval dataset in Table~\ref{tab:examples}. 
From this set, we sample from the 500 most frequent ($p_w$, $a_w$) pairs and randomly sample 100 unique context words $C$.~\footnote{We sample a limited number of context words to keep the scale of data annotation feasible.} Combining the sampled pun pairs and context words, we construct 4,552 ($C$, $p_w$, $a_w$) instances for annotation. %% ANJALI footnote and last sentence
% \anjali{[we need to justify here: why 100 context words? randomly sampled?]} 
% The prepared dataset contains , which results in an average of about 46 pun pairs per context word. Combining these results in a total} set of .
% Then, we extended the prepared dataset and finally got a total of set 4,552 (context word, $pw$, $aw$) pairs. The ($pw$, $aw$) pairs we prepare are sampled from the 500 most frequent ($pw$, $aw$) pairs in the SemEval dataset. The prepared dataset contains 100 unique context word, which results in an average of about 46 pun word pairs per context word. For our annotation, we aim to determine if users can come up with a pun using the pun word pair ($pw$, $aw$) that situates in a given context. If so, we ask the user to write down the pun.

\paragraph{Annotation.} For our annotation task, we asked annotators to indicate whether they can come up with a pun, using pun pair ($p_w$, $a_w$), that is situated in a given context $C$ and supports both senses $S_{p_w}$ and $S_{a_w}$. If an annotator indicated that they could create such a pun, we then asked the annotator to write down the pun they came up with. Meanwhile, we asked annotators how difficult it is for them to come up with the pun from a scale of 1 to 5, where 1 means very easy and 5 means very hard.~\footnote{Full annotation guidelines in Appendix~\ref{appendix-guidelines}.} To aid in writing puns, we also provided four T5-generated puns as references.~\footnote{Annotators find it extremely hard to come up with puns from scratch. Generated texts greatly ease the pain.} 

% Each HIT contains three ($C$, $p_w$, $a_w$) tuples and we pay one US dollar per HIT. 
% For each HIT, we provide context words $C$, the pun-alternative words $p_w$ and $a_w$, and word senses $S_{p_w}$, $S_{a_w}$ along with four generated puns as reference.~\footnote{Workers find it extremely hard to come up with puns from scratch. Generated texts greatly ease the pain.} 

% All annotators working on our HITs have collaborated with us in previous research projects and have been labeled as good annotators in the past. 

% During the annotation, we ask annotators to annotate if they can come up with a pun that situates in context words $C$, using pun words $pw$. At the same time, we require that both $S_{pw}$ and $S_{aw}$ should be supported in their puns.  

We deployed our annotation task on Amazon Mechanical Turk using a pool of 250 annotators with whom we have collaborated in the past, and have been previously identified as good annotators. Each HIT contained three ($C$, $p_w$, $a_w$) tuples and we paid one US dollar per HIT.\footnote{This translates to be well over \$15/hr.} 
%  from about 250 good MTurk workers who have collaborated with us in other projects} 
To ensure dataset quality, we manually checked the annotations and accepted HITs from annotators who tended not to skip all the annotations (i.e., did not mark everything as ``cannot come up with a pun''). After iterative communication and manual examination, we narrowed down and selected three annotators that we marked as highly creative to work on the annotation. To check inter-annotator agreement, we collected multiple annotations for 150 instances and measured agreement using Fleiss' kappa~\cite{Fleiss1973TheEO} ($\kappa=0.43$), suggesting moderate agreement.
% The IAA of binary deciding if a ($pw$, $aw$) pair is suitable for a given context to come up with a pun is 0.43, suggesting a moderate agreement.

\paragraph{Statistics.} After annotation, we ended up with 2,753 ($C$, $p_w$, $a_w$) tuples that are annotated as compatible and 1,798 as incompatible.
For the 2,753 compatible tuples, we additionally collected human-written puns from annotators. The number of puns we collected exceeds the number of annotated puns in SemEval 2017 Task 7 which have annotated pun word and alternative word sense annotations (2,396 puns). The binary compatibility labels and human-written puns comprise our resulting dataset, {\faMugHot} {\datasetname} (\textbf{C}ontext Sit\textbf{U}ated \textbf{P}uns). %% ANJALI number of puns
Table~\ref{tab:annotation} shows examples of annotations in \datasetname.

%% file: content/4-model.tex
\begin{figure*}
    \centering
    \includegraphics[width=\textwidth, trim={0 1cm 0 0.5cm},clip]{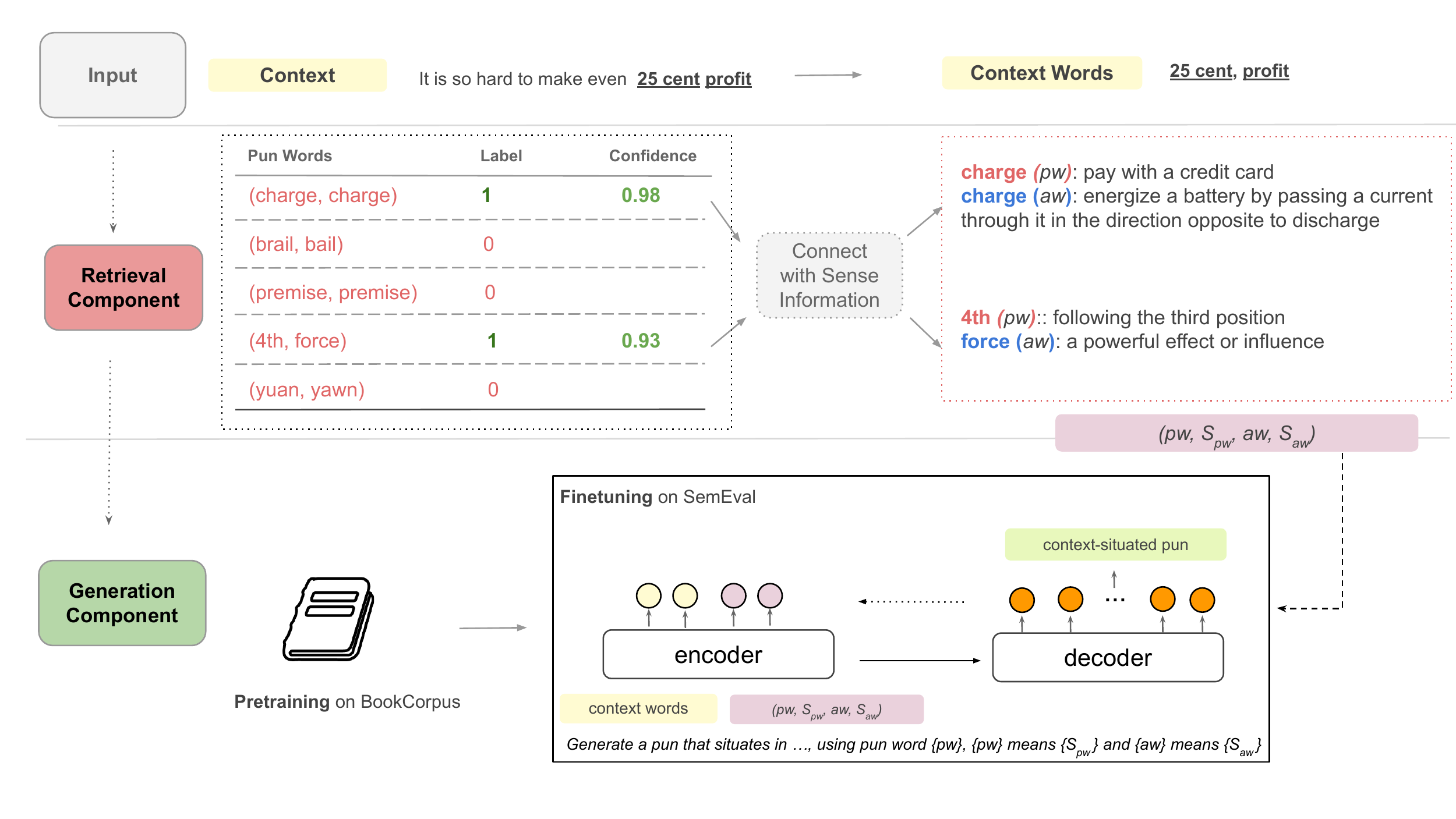}
    % \vspace{-0.25in}
    \caption{Our framework contains two components: (i) a retrieval component (top) that identifies relevant pun words for a given context, and (ii) a generation component (bottom) that takes the context and  retrieved pun words and generates context-situated puns.
    }
    \label{fig:system_design}
    % \vspace{-0.2in}
\end{figure*}

\section{Context-Situated Pun Generation}
%Given our dataset of puns, \datasetname, w
We propose a pipeline framework to generate context-situated puns, 
% with two components as shown in
shown in Figure~\ref{fig:system_design}. It consists of: (i) a retrieval-based module that selects a set of relevant pun word pairs,
% described in Section~\ref{sec:retrieval}, 
and (ii) a generation module that takes the context words and retrieved pun word pairs as input to generate puns. In this section, we briefly describe each component. % ANJALI
% described in Section~\ref{sec:pungen}.
% where we pretrain and then finetune an encoder-decoder architecture that leverages T5~\cite{t5}.

\paragraph{Pun Word Pair Retrieval.}\label{sec:retrieval}
We propose a retrieve-and-rank strategy to select $k$ relevant pun word pairs $(p_w, a_w)$ from a large, fixed set of pun word pairs $(P_{w}, A_{w})$ for a given context $C$. $C$ should be a list of keywords describing the context. If the context is given as a sentence, we use RAKE~\cite{rake} to automatically extract a list of keywords from the context to construct $C$.
For each context $C$, we apply a classifier to all available pun word pairs in our data $(P_w, A_w)$ and retrieve pairs classified as suitable. Then, we rank the suitable instances according to the model's confidence and take the top $k$ instances as the final retrieved $(p_w, a_w)$ pun word pairs. We experiment with both supervised and unsupervised approaches to build the retrieval module in Section~\ref{sec:pwpr}.  %% ANJALI

% \paragraph{Retrieval and Ranking.}
% \anjali{How is this different from the benchmark task? Can we refer to the benchmark task here (is it the same model)?} 
% \paragraph{Benchmark.}
% \anjali{this feels a bit out of place? is this just a side task?} 
% \datasetname serves as a good test bed for the binary task of deciding if a pun word pair $(p_w, a_w)$ is suitable for making a pun in a given context $C$.

% Table~\ref{tab:benchmark} shows the performance of finetuned pretrained language models on \datasetname. 

% then finetune pretrained NLI models. 
% We only keep ``entailment, contradiction'' labels for the NLI format. 

% First, we train a classifier to determine whether a $(p_w, a_w)$ is suitable for making a pun given $C$. 

% \paragraph{Modeling.} \anjali{[revise me?] We finetune a BERT\jiao{???} model~\cite{devlin-etal-2019-bert} for the classifier. Our input is the concatenated string \{$C$, $pw, aw$\}, and the output is the label 1 or 0 indicating the compatibility. We will justify our modeling choice through experiments later in \shereen{Section~\ref{sec:pwpr}}.}
\paragraph{Pun Generation.}
\label{sec:pungen}
Given pun word pair $(p_w, a_w)$, pun word senses $S_{p_w}$ and $S_{a_w}$, and context $C$, the pun generation module generates puns that relate to $C$, incorporate pun word $p_w$, and embody the meanings $S_{p_w}$ and $S_{a_w}$ of the pun word pair. Since there are limited pun datasets available for model training, we adopt a two-stage strategy that involves pretraining a T5-base~\cite{t5} model on non-pun text to learn to incorporate words and their senses in generations, then finetuning the model on pun data to learn the structure of puns. We describe our pun generation models in Section~\ref{sec:pun-gen-module}.
% To achieve this, we adopt a two-stage strategy including a pretraining stage and a finetuning stage. The target of pretraining is to let the model learn to incorporate the context words and pun word, and the target of finetuning is to further guide model to generate puns. The reason for adding the pretraining stage is because there are limited pun datasets available for model training. 
% For pretraining, we take the pretrained  as the base model.
%% ANJALI

%% file: content/5-experiment.tex
% We will discuss the experiments for both the retrieval component and the generation component separately in this section.

%retrieval_examples table was here, now in app

\section{Experiments}
We design our experiments to answer the following three research questions:
\paragraph{Q1.} What is the performance of the pun word pair retrieval module? (Section~\ref{sec:pwpr})
% Table~\ref{tab:retrieval_module} and Appendix \ref{app:retrieved_examples}, Table \ref{tab:retrieval_examples})
\paragraph{Q2.} What is the performance of the pun generation module? Is the pretraining stage necessary? (Section~\ref{sec:pun-gen-module})
% Table~\ref{tab:human_eval}-\ref{tab:puns})
\paragraph{Q3.} How well does the pipeline system perform in an end-to-end evaluation? Is the context-situated pun generation task plausible for humans? (Section~\ref{sec:e2e})
% , Table~\ref{tab:e2e}) 

% \begin{itemize}
%   \setlength\itemsep{0.1em}
%     \item \textbf{Q1}: How is the performance of the pun word pair retrieval module? (Section~\ref{sec:pwpr}, Table~\ref{tab:retrieval_examples}, Table~\ref{tab:retrieval_module})
%     \item \textbf{Q2}: How is the performance of the generation module? Is the pretraining stage necessary? (Section~\ref{sec:pungen}, Table~\ref{tab:human_eval}, Table~\ref{tab:puns})
%     \item \textbf{Q3}: How good the pipeline model is achieving compared to the human performance? (an end-to-end evaluation in Section~\ref{sec:e2e}, Table~\ref{tab:e2e}) 
% \end{itemize}

\subsection{Pun Word Pair Retrieval}
\label{sec:pwpr}
In this task, for a given context $C$ of keywords, the goal is to select $k$ relevant pun word pairs $(p_w, a_w)$ from a large, fixed set of pun word pairs $(P_{w}, A_{w})$.

\paragraph{Approaches.} We experiment with two approaches to building pun word pair retrieval systems, including supervised neural modeling and unsupervised embedding-based approaches.
% \begin{itemize}[leftmargin=*]
% \vspace{-0.1in}
% \itemsep-.3em 

    \medskip
    \noindent \emph{\underline{Neural}.} We finetune BERT-base~\cite{devlin-etal-2019-bert}, RoBERTa-base~\cite{liu2019roberta} and DeBERTa-base~\cite{he2021deberta} models on the CUP dataset for pun word pair classification. The input is formatted as sentence matching, where, given the context $C$ as sentence 1 and the pun word pair as sentence 2, the output label indicates if the two sentences are compatible. Additionally, we experiment with finetuning natural language inference (NLI) models, RoBERTa-large-NLI~\cite{liu2019roberta} and BART-large-MNLI~\cite{lewis-etal-2020-bart}. We use the context words as premise and the pun word pair as hypothesis, with \emph{entailment} and \emph{contradiction} labels as outputs. For each context, we retrieve all pun pairs classified as suitable by the model, then rank the instances according to the model's confidence
    (i.e., output from the last layer after softmax) to retrieve the top-$k$ pun pairs.
    
    \medskip \noindent \emph{\underline{Unsupervised}.} The key idea behind the compatibility classification is to find pun word pairs that are semantically close to the context. Therefore, a natural question to ask is, ``Can an unsupervised method that measures semantic similarity can perform as well as the neural method?'' Here, we use Euclidean distance between Glove embeddings~\cite{pennington-etal-2014-glove} of pun and context words to measure the semantic similarity. Formally, for a context $C$ consisting of a list of context words $c_1, c_2, ..., c_n$, we calculate the average Euclidean distance between the Glove  representation of $p_w, a_w$ and the embedding of each of context word $c_i$:
\begin{equation}
    \sum_{i=1}^{n} d(\vec{p_w}, \vec{c_i}) + \sum_{i=1}^{n} d(\vec{a_w}, \vec{c_i}).
\end{equation}
Then we rank all 500 possible $(p_w, a_w)$ candidates using the distance score above, retrieving the $k$ pairs with the smallest distance as the top-$k$ retrieved pun word pairs. %% ANJALI
% \end{itemize}

\begin{table}[t]
\small
\centering
\begin{tabular}{l|llll}
\toprule
 & \textbf{train} & \textbf{dev} & \textbf{test} & \textbf{total} \\ \midrule
\textbf{pos} & 1,873 & 290 & 590 & 2,753 \\ \midrule
\textbf{neg} & 1,282 & 175 & 341 & 1,798 \\ \midrule
\textbf{all} & 3,155 & 465 & 931 & 4,551 \\ \bottomrule
\end{tabular}
\caption{{\faMugHot} {\datasetname} data splits for the pun word pair retrieval task. We show the distribution of $(C, p_w, a_w)$ tuples labeled as suitable or unsuitable in each split.}
\label{tab:statistics_collect}
\end{table}

\begin{table*}[]
\small
\centering
\renewcommand{\tabcolsep}{1.2mm}
%\resizebox{\textwidth}{!}{
\begin{tabular}{@{}l|llll|llll@{}}
\toprule
 & \multicolumn{4}{c|}{dev} & \multicolumn{4}{c}{test} \\ \midrule
 & \textbf{F1} & \textbf{Precision} & \textbf{Recall} & \textbf{Acc} & \textbf{F1} & \textbf{Precision} & \textbf{Recall} & \textbf{Acc} \\ \midrule
bert-base~\cite{devlin-etal-2019-bert} & 62.29$_{1.18}$ & 62.23$_{1.16}$ & 62.58$_{1.28}$ & 64.02$_{0.96}$ & 62.39$_{0.34}$ & 62.30$_{0.31}$ & 62.56$_{0.38}$ & 64.70$_{0.28}$ \\
roberta-base~\cite{liu2019roberta} & 63.91$_{0.79}$ & 63.88$_{0.72}$ & 64.57$_{0.71}$ & 65.09$_{0.90}$ & 61.85$_{0.17}$ & 61.73$_{0.17}$ & 62.14$_{0.09}$ & 63.91$_{0.32}$ \\
deberta-base~\cite{he2021deberta} & 63.93$_{0.62}$ & 63.84$_{0.58}$ & 64.14$_{0.70}$ & 65.73$_{0.54}$ & 62.55$_{1.49}$ & 62.48$_{1.47}$ & 62.70$_{1.59}$ & 64.91$_{1.28}$ \\ \midrule
roberta-large-nli~\cite{liu2019roberta} & 67.25$_{0.69}$ & 67.13$_{0.70}$ & 67.45$_{0.65}$ & 68.96$_{0.71}$ & 64.72$_{0.42}$ & 64.96$_{0.63}$ & 64.60$_{0.30}$ & 67.60$_{0.73}$ \\
bart-large-nli~\cite{lewis-etal-2020-bart} & 67.33$_{0.74}$ & 67.28$_{0.82}$ & 67.54$_{0.52}$ & 69.03$_{1.05}$ & 63.81$_{0.39}$ & 63.83$_{0.53}$ & 63.87$_{0.20}$ & 66.31$_{0.79}$ \\ \bottomrule
\end{tabular}%}
\caption{Pun word classification performance of neural models on \datasetname, showing that our task is challenging for pretrained LMs. We report models' performance across three random seeds with standard deviation as subscripts.}
\label{tab:benchmark}
\end{table*}

\begin{table}[]
\small
\resizebox{\columnwidth}{!}{
\begin{tabular}{@{}ccccc}
\toprule
& \textbf{TP@1} & \textbf{TP@5} & \textbf{TP@10} & \textbf{TP@20} \\ \midrule
Unsupervised        & 64.0              & 59.4                 & 60.2              & \textbf{61.5}             \\ \midrule
\textbf{{\small\faTrophy} Neural} & \textbf{69.0}             & \textbf{63.2}             & \textbf{61.7}              & 59.3             \\ \bottomrule
\end{tabular}
}
\caption{TP@N results for supervised (neural) and unsupervised approaches for pun word retrieval. \texttt{TP} stands for True Positive rates.}
\label{tab:retrieval_module}
\end{table}

\paragraph{Experiment Setup.}  We split \datasetname into 70\% training, 10\% validation and 20\% test data.
Table~\ref{tab:statistics_collect} shows the distribution of pun word compatibility labels in our data splits. For each context word, we use our models to retrieve pun word pairs from 500 candidate pairs for making context-situated puns.~\footnote{Further experimental details in Appendix~\ref{sec:appendix-classifier}.}

% In addition to our neural method described in , we  modeling strategies: neural and unsupervised methods 

\paragraph{Evaluation Metrics.} 
% We report \emph{HIT@N (\%)} as the evaluation metric to evaluate the quality of the retrieval module. More specifically, we report HIT@1, HIT@5, HIT@10, HIT@20 in our work. HIT@N measures if the ground-truth pun word pair is within the top $N$ retrieved pun word pairs. Then \emph{HIT@N (\%)} is the percent of HIT@N among all test instances.  For example, if the pun word pair is on the first position in the predicted pun word list, then we add 1 to HIT@1, HIT@5, HIT@10 and HIT@20. 
For neural models, we first benchmark the accuracy, precision, recall, and F1 of the model's predictions for the pun word pair classification task on the CUP dataset. Additionally, for both approaches, we use the True Positive rate (TP@N) to evaluate the performance of our pun word retrieval module. It measures the percentage of top-$k$ retrieved pun word pairs that can be used to generate puns for a given context. The higher the TP@N is, the stronger the retrieval module is in terms of retrieving appropriate pun word pairs. %% ANJALI

\paragraph{Results.} We show results of our supervised pun word classifiers in Table~\ref{tab:benchmark}. Our results show that the task of classifying whether a context word is compatible with a pun word pair is challenging for current pretrained LMs, with a best F1 of 64.72 from RoBERTa-large-NLI. 
Table~\ref{tab:retrieval_module} shows the TP@N evaluation of pun word pairs retrieved by our best neural model, finetuned RoBERTa-large-NLI, and our unsupervised method. In general, the supervised neural model outperforms the unsupervised method. TP@1 shows that 69\% of pun word pairs retrieved by the neural model are compatible with their given context, showcasing the effectiveness of our retrieval module. We provide additional qualitative analysis in Appendix \ref{app:retrieved_examples}, Table~\ref{tab:retrieval_examples}. %% ANJALI

\subsection{Pun Generation}
\label{sec:pun-gen-module}

Given pun word pair $(p_w, a_w)$, pun word senses $S_{p_w}$ and $S_{a_w}$, and context $C$, the task is to generate a pun that relates to $C$, incorporates pun word $p_w$, and utilizes both pun word senses $S_{p_w}$ and $S_{a_w}$. %% ANJALI

\paragraph{Approach.} %We compare our generation module performance with its variant which takes out the pretraining component.
For the novel task of context-situated pun generation, we establish a baseline model that uses a combination of pretraining on non-pun text and finetuning on pun text to generate both homographic and heterographic puns. Our unified framework for homographic and heterographic pun generation is also new to the community. We evaluate the following model variants: %% ANJALI unified framework new

\medskip 
% For the baseline method, as the task is new and there have not been any works on either context-situated pun generation. We compare our model with its variant where we take out the pretraining stage and only keep the finetuning stage. 
\noindent \ul{\textit{AmbiPun}}~\citep{mittal2022ambipun}. Previous systems for heterographic pun generation explicitly require homophones, making it hard to adapt them to homographic puns~\cite{yu-etal-2020-homophonic, he-etal-2019-pun}. Therefore, we use AmbiPun, a the state-of-the-art homographic pun generation model, to generate both homographic and heterographic puns without further finetuning. Following their prompt format, we use ``generate sentence: \{\textrm{$C$}\}, \{\textrm{$p_w$}\}, \{\textrm{$a_w$}\}'' for homographic puns and ``generate sentence: \{\textrm{$C$}\}, \{\textrm{$p_w$}\}'' for heterographic puns. 

\medskip
\noindent \ul{\textit{Finetuned T5}} (T5$_\textrm{FT}$). We finetune T5-base~\cite{t5} on the SemEval 2017 Task 7 dataset~\cite{miller-etal-2017-semeval}, in which puns are annotated with pun word pairs $p_w$ and $a_w$ along with their sense information $S_{p_w}$ and $S_{a_w}$. We construct $C$ using the RAKE~\cite{rake} keyword extraction algorithm on the pun text, and further verify them against human-annotated keywords from an augmentation of the SemEval dataset we designed to enable keyword-conditioned pun generation~\cite{sun2022expun}.
% ~\footnote{The extracted keywords are further confirmed by human-annotated keywords through crowdsourcing. The dataset effort is in parallel submission with this paper. We will cite that paper in the camera-ready version.} 
During finetuning, we use the input prompt: ``\emph{generate a {pun} that situates in} \{\texttt{$C$}\}, \emph{using the word} \{\texttt{$p_w$}\}, \{\texttt{$p_w$}\} \emph{means} \{\texttt{$S_{p_w}$}\} \emph{and} {\{\texttt{$a_w$}\} \emph{means} \{\texttt{$S_{a_w}$}\}}''. 
% We use underline to highlight the difference in input prompt between the pretraining and finetuning stages.
% as input, where we use underline to highlight the difference between the finetuning and pretraining inputs. 
% Another difference between the finetuning and pretraining is that, during the keyword extraction process, we do not filter out keywords from RAKE output. 
The goal of finetuning is to teach the model to incorporate both word senses in the final generated puns. 

\medskip
\noindent \ul{\textit{Finetuned T5 with pretraining}} (T5$_\textrm{PT+FT}$). Here, we investigate whether the model can learn to incorporate words and their senses into the generated sentences by pretraining on non-pun text. To this end, we automatically construct a pretraining dataset from BookCorpus~\cite{bookcorpus}. For each word $w\in \{p_w, a_w\}$ in a given pun word pair,~\footnote{We select heterographic pun pairs ($p_w \neq a_w$) to avoid introducing polysemic ambiguity in the pretraining stage.}  
% At the same time, we also know their sense information $S_{pw}$ or $S_{aw}$. 
we mine 200 sentences that contain $w$ from BookCorpus.~\footnote{We lemmatize $w$ and the sentence using Spacy (\url{https://spacy.io/}) so grammatical features will not have impact on our mining.} We extract keywords from a given BookCorpus sentence containing $w$ using RAKE to construct context $C$. We retain noun and verb keywords,
% , specifically, any ``NOUN'' or ``VERB'' keywords,
as they are more likely to have significant impact at the sentence level~\cite{kim2000patterns, cutler1977role}, and exclude pun word $w$ from the keyword list. Using these automatically-constructed samples, we finetune T5~\cite{t5} to generate sentences situated in $C$ that incorporate $w$, using the input prompt: ``\emph{generate a sentence that situates in} \{\texttt{$C$}\}, \emph{using the word} \{\texttt{$w$}\}, \{\texttt{$w$}\} \emph{means} \{\texttt{$S_{w}$}\} \emph{and} \{\texttt{$w$}\} \emph{means} \{\texttt{$S_{w}$}\}'', the output of which is the retrieved sentence from BookCorpus that uses $C$ and $w$. 
% We repeat the process for $aw$ by replacing all information for $pw$ with $aw$'s. 

\paragraph{Experiment Setup.} We finetune our T5 models on 1,382 training samples from SemEval Task 7 that contain both pun word and sense annotations. For testing, we randomly sample 200 ($C$, $p_w$, $a_w$) tuples from CUP that annotators marked as compatible. We use each model to generate puns for this set and compare their performance.~\footnote{Further experimental details in Appendix~\ref{sec:appendix-t5}.}

\begin{table}[t]
\small\centering
\begin{tabular}{lcc}
\toprule
\textbf{Model} & \textbf{$p_w$ Incorp. \%} & \multicolumn{1}{c}{\textbf{Success \%}} \\ \midrule
AmbiPun & \textbf{97.22} & 11.11 \\ \midrule
T5$_\textrm{FT}$ & 96.67 & 23.89 \\ \midrule
\textbf{{\small\faTrophy}} \textbf{T5$_\textrm{PT+FT}$} & \textbf{97.22} & \textbf{31.11} \\ \bottomrule
\end{tabular}
\caption{Pun generation results using automatic (pun word incorporation) and human (success rate) evaluation. We compare our finetuned T5 models to a state-of-the-art baseline, AmbiPun~\cite{mittal2022ambipun}. \texttt{PT} stands for Pre-Training and \texttt{FT} stands for Fine-Tuning.}
% The performance comparison on randomly-sampled 200 (context word, pun word pair) combinations between our generation module, its variant without pretraining and the current state-of-the-art AmbiPun~\cite{mittal2022ambipun}. Our generation module yields the best performance, with a success rate almost tripling AmbiPun.}
\label{tab:human_eval}
\end{table}

\begin{table*}[]
\small\centering
\renewcommand{\tabcolsep}{1.5mm}
\begin{tabular}{llll}
\toprule
\textbf{Context} & $p_w$/$a_w$ & \multicolumn{2}{l}{\textbf{Generated Pun}} \\ \midrule
scientist, liquid & assay/ & \textit{Ours:} & A scientist who is a liquid chemical expert can't assay the problem. \\ 
chemicals, problem & say & \textit{Ambi.:} & What do you call a scientist with a liquid chemicals problem? an assay-ist. \\ \midrule
fruit vendor, & yammered/ & \textit{Ours:} & She was only a Fruit Vendor's daughter, but she yammered. \\ 
 daughter & yam &  \textit{Ambi.:} & My daughter yammered at the fruit vender... she said i'm not a fruit vender.\\ \midrule
{\multirowcell{3}[0pt][l]{opera, orchestra \\ conductors}} & \multirowcell{3}[0pt][l]{pitch/ \\ pitch} & \textit{Ours:} & Conductors of the opera had to make a good pitch. \\ 
  & & \textit{Ambi.:} &  Why do opera and orchestra conductors pitch their voices? \\ 
  & & & because they can't sing. \\ \midrule
\multirowcell{3}[0pt][l]{company football \\ team, meeting, \\ get together} & \multirowcell{3}[0pt][l]{kickoff/ \\ kickoff} & \textit{Ours:} & A football team's meeting was about to kick off. \\ 
 &  &  \textit{Ambi.:} & I'm going to get together for a company football team meeting \\
 & & & before kickoff. \\ \bottomrule
\end{tabular}
\caption{Examples of generated context-situated puns from our system and AmbiPun~\cite{mittal2022ambipun}.}
\label{tab:puns}
\end{table*}

\paragraph{Evaluation Metrics.} We report \emph{pun word incorporation rate} as the automatic evaluation metric to measure the model's ability to incorporate pun words in the final generation. We also conduct human evaluation on Amazon Mechanical Turk to judge whether the generated puns are successful.~\footnote{Turkers had to pass a qualifier by correctly labeling $>=80\%$ of 20 samples that we manually annotated. Success is defined as whether the text supports both senses of the pun word. We measure inter-annotator agreement among 3 annotators using Fleiss' kappa ($\kappa=0.49$), showing moderate agreement.}
% We set a qualification test that contain 20 puns to filter out workers that do not understand the definition of puns. For the qualification, we keep the ones that score higher than 80\% to work on our evaluation task. One pun instance has three annotators, among whom we take the majority as the final judgement. The fleiss kappa score among three annotators are 0.49, suggesting a moderate agreement.

\paragraph{Results.} Pun generation results are shown in Table~\ref{tab:human_eval}. We find that: (1) adding the pretraining stage helps our model better incorporate pun words, and (2) our generation module can generate successful puns at almost triple the rate of the current state-of-the-art framework AmbiPun (examples in Table~\ref{tab:puns}). We hypothesize that this is because AmbiPun is a completely unsupervised approach in which the pun generator is not finetuned on any pun data, and because our models additionally benefit from rich word sense information in the input.
% {\color{red} this is probably mainly because we have sense annotations in the input, though...? needs more discussion?}

% \begin{table*}[]
% \small\centering
% \begin{tabular}{l|l|l|l}
% \hline
% \textbf{Context} & \bf pw/aw & \textbf{Model} & \multicolumn{1}{c}{\textbf{Generation}} \\ \hline
% scientist, liquid & assay & Ours & A scientist who is a liquid chemical expert can't assay the problem. \\  
%   chemicals, problem & say &  Ambi. & What do you call a scientist with a liquid chemicals problem? an assay-ist. \\ \hline
% fruit vendor, & yammered & Ours & She was only a Fruit Vendor's daughter, but she yammered. \\ 
%  daughter & yam &  Ambi. & My daughter yammered at the fruit vender... she said i'm not a fruit vender.\\ \hline
% opera, & pitch& Ours & Conductors of the opera had to make a good pitch. \\ 
%  orchestra conductors & pitch &  Ambi. & Why do opera and orchestra conductors pitch their voices? because they can't sing. \\ \hline
% company football team, & kickoff & Ours & A football team's meeting was about to kick off. \\ 
%  meeting, get together & kickoff &  Ambi. & I'm going to get together for a company football team meeting before kickoff. \\ \hline
% \end{tabular}
% \caption{\shereen{Examples of generated context-situated puns from our system and AmbiPun~\cite{mittal2022ambipun}.}}
% \label{tab:puns}
% \end{table*}

\input{content/5-0-e2e}

%% file: content/5-0-e2e.tex
\subsection{End-to-end Evaluation}
\label{sec:e2e}

Finally, we evaluate how well our pipeline retrieves relevant pun word pairs and generates novel puns given a context of keywords in an end-to-end fashion, and compare our pipeline's performance to human-standard annotations from CUP.

\begin{table}[t]
\small\centering
\renewcommand{\tabcolsep}{1.8mm}
\begin{tabular}{@{ }clccc@{}}
\toprule
\textbf{Retrieval} & \multicolumn{1}{c}{\textbf{Generation}} & \multicolumn{2}{c}{\textbf{Incorp. \%}} & \textbf{Success \%} \\
 Sec~\ref{sec:pwpr} & \multicolumn{1}{c}{Sec~\ref{sec:pun-gen-module}} & $C$ & $p_w$ &  \\ \midrule
% \begin{tabular}[c]{@{}c@{}}\textbf{Retrieval} \\ (5.1) \end{tabular} & \begin{tabular}[c]{@{}c@{}}\textbf{Generation} \\ (5.2)\end{tabular} & \multicolumn{1}{c}{\textbf{$C$}} & \multicolumn{1}{c}{\textbf{$p_w$}} & \multicolumn{1}{l}{\textbf{Success \%}} \\ \midrule
\multicolumn{2}{c}{{Human}} & 81.94 & 75.67 & \textbf{71.67} \\ \midrule  %\small\faTrophy} {\small\faTrophy} 
\multicolumn{1}{l}{\multirow{3}{*}{Unsup.}} & AmbiPun & \textbf{100.00} & \textbf{92.66} & 10.00 \\ %\cmidrule{2-5}
\multicolumn{1}{l}{} & T5$_\textrm{FT}$ & 91.67 & 80.76 & 26.67 \\ %\cmidrule{2-5}
\multicolumn{1}{l}{} & T5$_\textrm{PT+FT}$ & 97.22 & 80.74 & 26.67 \\ \midrule
\multicolumn{1}{l}{\multirow{3}{*}{Neural}} & AmbiPun & 98.51 & 97.34 & 21.67 \\ %\cmidrule{2-5}
\multicolumn{1}{l}{} & T5$_\textrm{FT}$ & 91.04 & 78.08 & 23.33 \\ %\cmidrule{2-5}
\multicolumn{1}{l}{} & T5$_\textrm{PT+FT}$ & 97.01 & 79.83 & \textbf{40.00} \\ \bottomrule %{\small\faTrophy} 
\end{tabular}
\caption{End-to-end evaluation of our system against AmbiPun and human baselines. }
\label{tab:e2e}
\end{table}

 \paragraph{Experiment Setup.} We randomly choose 60 context words to conduct the end-to-end evaluation. For each context, we use both unsupervised and neural pun word retrieval modules from Section~\ref{sec:pwpr} to retrieve the \emph{top-1} predicted pun word pair, then use each of the pun generation modules from Section~\ref{sec:pun-gen-module} to generate puns using the retrieved pun word pair. We also compare with human performance. For each context, we find the human-written pun in CUP that annotators indicated was least difficult to write, randomly sampling one pun in case of ties. We use annotation difficulty as a proxy for ranking human context-situated puns, assuming more natural puns are easier to write.
 
\paragraph{Evaluation Metrics.} We measure the incorporation rate of context words $C$ and pun words $p_w$ as automatic evaluation metrics. In addition, similar to standalone pun generation evaluation, we conduct human evaluation to judge whether the generated puns are successful.

\paragraph{Results.} We report results of combinations of our retrieval and generation modules in Table~\ref{tab:e2e}.
%evaluation of the combination of our best performed module for retrieval and generation in Table~\ref{tab:e2e}.
We show that: (i) our pretraining step is helpful in terms of both improving the keyword incorporation rate and pun success rate of the generation module, despite using retrieved pun words as input. (ii) Our pipeline system performs the best among all model variations, yielding a success rate of pun generation of 40\%. This success rate improves over the best reported in Section~\ref{sec:pun-gen-module} (31\%), showcasing the benefit of using our neural pun word retrieval module over randomly sampling pun word pairs for a given context. However, (iii) the best model performance is still about 32\% lower than the human success rate, indicating that humans can complete the context-situated pun generation task plausibly and much more successfully, indicating large room for improvement.
% The higher success rate than Table~\ref{tab:human_eval}'s where we do randomly sampling instead of top-1 retrieved pun word pair also showcases the effectiveness of our pun word retrieval module.  However, (iii) the best model performance is still about 32\% lower than the human success rate, indicating the plausibility of the challenging context-situated pun generation task. 

%% file: content/related.tex
\section{Related Work}
Our work proposes an approach for conditional generation of humor using a retrieve-and-generate framework. More specifically, our work enables a constrained type of pun generation. We briefly summarize existing work in these directions.
% Our work is a constrained type of \emph{pun generation} and proposes to a framework of \emph{retrieve and generate}. From a broader aspect, it is related to \emph{conditional generation} and \emph{humor generation}. We briefly summarize work in these directions.
% \paragraph{Conditional generation.}
% Context-situated pun generation falls broadly into the category of conditional generation. While previous work on pun generation can also be seen as conditional generation where the generated puns should contain the given pun words, our work adds another constraint that the generated puns should also be related to given context. Other conditional generation works constrain on syntactic information (e.g., paraphrase generation~\cite{sun-etal-2021-aesop, huang-chang-2021-generating}), rhythmic information (e.g., sonnet generation~\cite{tian2022zero}), and structural information (e.g., event extraction by generation~\cite{hsu2022degree}).

\paragraph{Humor generation.} %A major challenge for research in computational humor has been lack of datasets.  
With the recent advent of diverse datasets \cite{hasan2019ur,mittal2021so,yang-etal-2021-choral}, it has become easier to detect and generate humor. While large pretrained models have been fairly successful at humor detection, humor generation still remains an unsolved problem, and is usually studied in specific settings.~\citet{petrovic-matthews-2013-unsupervised} generate jokes of the type `I like my X like I like my Y, Z'. \citet{garimella-etal-2020-judge} develop a model to fill blanks in a Mad Libs format to generate humorous sentences, and \citet{yang-etal-2020-textbrewer} edit headlines to make them funny. More research is required to generate humorous sentences that are not constrained by semantic structure.

\paragraph{Retrieve and generate.}
Our work proposes a retrieval and generation pipeline for generating context-situated puns. The retrieval component finds proper pun word pairs for the current context, and the generation component generates puns utilizing context words and pun word pairs. Similarly, ~\citet{yu-etal-2020-homophonic} adopt a pair of homophones, retrieve sentences that contain either word from a large corpus then edit the sentence into a pun sentence. ~\citet{sun-etal-2021-aesop} first retrieve syntactic parses and then generate paraphrases that keep the semantic meaning while conforming to the retrieved syntactic parses. 

\paragraph{Pun generation.}
Previous work on pun generation has focused on heterographic pun generation or homographic pun generation \cite{miller-gurevych-2015-automatic, hong-ong-2009-automatically, petrovic-matthews-2013-unsupervised,inproceedings2000}. At the same time, all of them require an input of pun words and assume pun words are given. Heterographic pun generation requires a pair of homophones, and homographic pun generation requires a polyseme, i.e., a pun word that has more than one meaning.  \citet{he-etal-2019-pun} make use of local-global surprisal principle to generate heterographic puns and \citet{yu-etal-2020-homophonic} use constrained lexical rewriting for the same task. \citet{hashimoto2018retrieve} use a retrieve and edit approach to generate homographic puns and \citet{yu-etal-2018-neural,luo-etal-2019-pun} propose complex neural model architectures such as constrained language model and GAN. ~\citet{mittal2022ambipun} generate homographic puns given a polyseme and try to incorporate the multiple senses of the polyseme. 
~\citet{tian2022unified} proposed a unified framework to generate both homographic and homophonic puns. 
Our setting is different from all previous work, first asking what pun words we should use for generating a pun in a given context. Meanwhile, our work finds the connection between heterographic pun generation and homographic pun generation: both types must utilize the two meanings of a pair of words. For heterographic pun generation, the two meanings come from the pair of homophones, while for homographic pun generation, the two meanings come from the polyseme itself.  Motivated by this, we propose a unified framework that can generate both heterographic puns and homographic puns adaptively.

%% file: content/99-conclusion.tex
\section{Conclusion and Future Work}
We propose a new setting for pun generation: \textit{context-situated pun generation}.
As a pioneering work, to facilitate future research in this direction, we first collect a large-scale corpus, {\faMugHot}~\datasetname,  which contains 4,551 annotated context and pun word pairs annotated for compatibility, along with 2,753 human-written puns for the compatible pairs, which is of an even larger size than the current most commonly-used pun dataset, SemEval 2017 Task 7~\cite{miller-etal-2017-semeval}. 
To benchmark the performance of the state-of-the-art NLG techniques on the proposed task, we build a pipeline system composed of a pun pair retrieval module that identifies suitable pun pairs for a given context, and a generation module that generates context-situated puns given the context and compatible pun pairs. 
Human evaluation shows that the best model achieves 40\% success rate in end-to-end evaluation, trailing behind human performance by almost 32\%, highlighting the challenge of the task and encouraging more future work in this direction.  %69\% of our top retrieved pun words can be used to generate puns. 

% For the pun generation module, we propose a pretraining and finetuning framework. 
% Our generation module is the first effort that unifies the homographic pun generation and heterographic pun generation in the pun generation domain. 
% In addition, it surpasses the current state-of-the-art generation model in terms of successfully generating puns.
Our work introduces the concept of \emph{situating in context} to pun generation. However, future work can easily extend the idea and framework to other areas of creative generation, such as metaphor generation, lyric generation, and others. %Considering the context while generating will make the generation more engaging and interesting. 
Another promising future direction is to integrate the generated puns into the original conversational or situational context to improve the interestingness and engagingness of the downstream applications. 
We hope our work can inspire more innovations on context-situated creative generation.

%% file: content/limitations.tex
\section*{Acknowledgements}
We thank PlusLab members for the discussion and early feedback about this new setup. We also thank anonymous reviewers for their constructive feedback and suggestions that helped improve the draft.

\section*{Limitations}
In this work, we focus on the task of pun generation, a specific area of creative language and humor generation. We acknowledge that humor is a highly subjective area, i.e., what might be perceived as humorous may differ greatly from one person to another depending on their unique backgrounds and experiences. We hope this work and dataset can be used more broadly to give insight into how humor can differ based on contextual nuances and personal characterizations.

\section*{Ethics}
We hereby acknowledge that all of the co-authors of this work are aware of the provided \textit{ACL Code of Ethics} and honor the code of conduct.

Since we use pretrained language models for our generation tasks, we note that this makes our models susceptible to generating biased or sensitive content. While we do not explicitly address concerns around bias/sensitive content within our framework to date, we aim to incorporate these considerations into pun generation as we develop new models, including methods to filter our inputs and generated data for toxicity and biased references that may be deemed offensive.

%% file: appendix/modeling_details.tex
\section{Classifier Implementation Details}~\label{sec:appendix-classifier}
We finetune five pretrained language models for classifying whether context words and pun word pairs are compatible in Table~\ref{tab:benchmark}, and we use HuggingFace~\cite{wolf-etal-2020-transformers} throughout our implementation for accessing model checkpoints and modeling. For hyper-parameter search, we tried the combinations of learning rate \{$1e^{-4}$, $3e^{-4}$, $1e^{-5}$, $3e^{-5}$\} * training epoch \{3, 10, 20\}. The final hyperparameters for bert-base, roberta-base and deberta-base are: learning rate $1e^{-5}$, training epoch 20 and training batch size 32. For roberta-large-mnli and bart-large-mnli models, we reduce the training epochs to 3 and training batch size to 8. We choose the checkpoint with the best accuracy on the dev set for inference.

%Hi anjali, elv model doesn't need to include the description for the mnli models and below is details for ELV model.
%   --labeled_data ${LABELED_DATA} --unlabeled_data ${UNLABELED_DATA} \
%   --valid_data ${VALID_DATA} --test_data ${TEST_DATA} \
%   --num_labels 3 --output_dir ${OUTPUT_DIR} \
%   --do_lower_case --max_source_seq_length 128 --max_target_seq_length 128 \
%   --num_classifier_epochs_per_iters 3 --num_generator_epochs_per_iters 5 \
%   --per_gpu_train_batch_size 4 --gradient_accumulation_steps 4 \
%   --num_iters 20 --learning_rate 1e-5 --num_warmup_steps 500 --cache_dir ${CACHE_DIR}
% ------------------ uncomment from here -----------------------
% For ELV model, we use the released code  and inherented most of their default hyperparameters for \emph{ELV-sa}\footnote{\url{https://github.com/JamesHujy/ELV/blob/main/EST_sa/train_restaurant.sh}}. We changes the number of training batch size per GPU to 4 to accelerate the training.

\section{T5 Implementation Details}
\label{sec:appendix-t5}
We finetune multiple T5 models~\cite{t5} in our work, and we use T5-base from SimpleT5~\footnote{\url{https://github.com/Shivanandroy/simpleT5}} throughout our implementation. We use 512 and 256 for the maximum source length and the maximum target length respectively.
As the optimizer, we use AdamW~\cite{adamw} with a learning rate of 0.0001.
For the pretraining stage, we finetune T5 for 3 epochs on retrieved BookCorpus data. During the finetuning stage, we train each model on a Tesla V100 with a batch size of 8 for 30 epochs. 
During inference, we use beam search as the decoding method with a beam size of 2. 
We terminate decoding when the EOS token is generated or the maximum target length is reached.

%% file: appendix/retrieval_examples.tex
\section{Retrieved Pun Word Pair Examples}
\label{app:retrieved_examples}

\begin{table*}[t!]
\small
\resizebox{\textwidth}{!}{
\begin{tabular}{@{}l|l|l|l@{}}
\toprule
\textbf{Context} &
  \textbf{SemEval Annot.} &
  \textbf{Modeling} &
  \textbf{Retrieved Pun Word Pairs} \\ \midrule
\multirow{2}{*}{einstein, parents} &
  \multirow{2}{*}{relatively, relativity} &
  Unsupervised &
  (kid, kid), (father, feather), (allow, aloud), (census, sense), (throng, wrong) \\ \cmidrule(l){3-4} 
 &
   &
  neural &
  \textbf{(relatively, relativity)}, (pinch, pinch), (pupil, pupil), (father, feather), (kid, kid) \\ \midrule
\multirow{2}{*}{bright star} &
  \multirow{2}{*}{seriously, sirius} &
  Unsupervised &
  (bright, bright), (guess, guest), (limelight, lime), (father, feather), (mist, miss) \\ \cmidrule(l){3-4} 
 &
   &
  neural &
  \begin{tabular}[c]{@{}l@{}}(bright, bright), (light, light), (constellation, consolation), \textbf{(seriously, sirius)},\\ (serious, sirius)\end{tabular} \\ \midrule
\multirow{2}{*}{interpreters, die} &
  \multirow{2}{*}{sign, sign} &
  Unsupervised &
  (go, go), (turn, turn), (dye, die), (throng, wrong), (get, get) \\ \cmidrule(l){3-4} 
 &
   &
  neural &
  \begin{tabular}[c]{@{}l@{}}(connection, connection), (dye, die), \textbf{(sign, sign)}, (sentence, sentence),\\ (fluently, flue)\end{tabular} \\ \bottomrule
\end{tabular}
}
\caption{Examples of retrieved pun word pairs from both the \emph{unsupervised} and the \emph{neural} method. We highlight the annotated pun word pairs from the SemEval dataset in the prediction list in \textbf{bold}. However, using the annotated pun word pairs as the only ground truth underestimates the pun word retrieval module. As shown here, both methods can retrieve pun word pairs that are related to the context. However, these (context words, $p_w$, $a_w$) combinations are missing from the original SemEval annotations. This again highlights the importance of collecting {\faMugHot} {\datasetname} that includes (context words, $p_w$, $a_w$) pairs to facilitate future studies in the context-situated pun generation domain.}
\label{tab:retrieval_examples}
\end{table*}

Table \ref{tab:retrieval_examples} shows examples of retrieved pun word pairs from both the unsupervised and neural methods.

%% file: appendix/annotation-guideline.tex
\section{Annotation Guidelines}\label{appendix-guidelines}
 \begin{figure*}[h!]
    \centering
    \includegraphics[width=\textwidth]{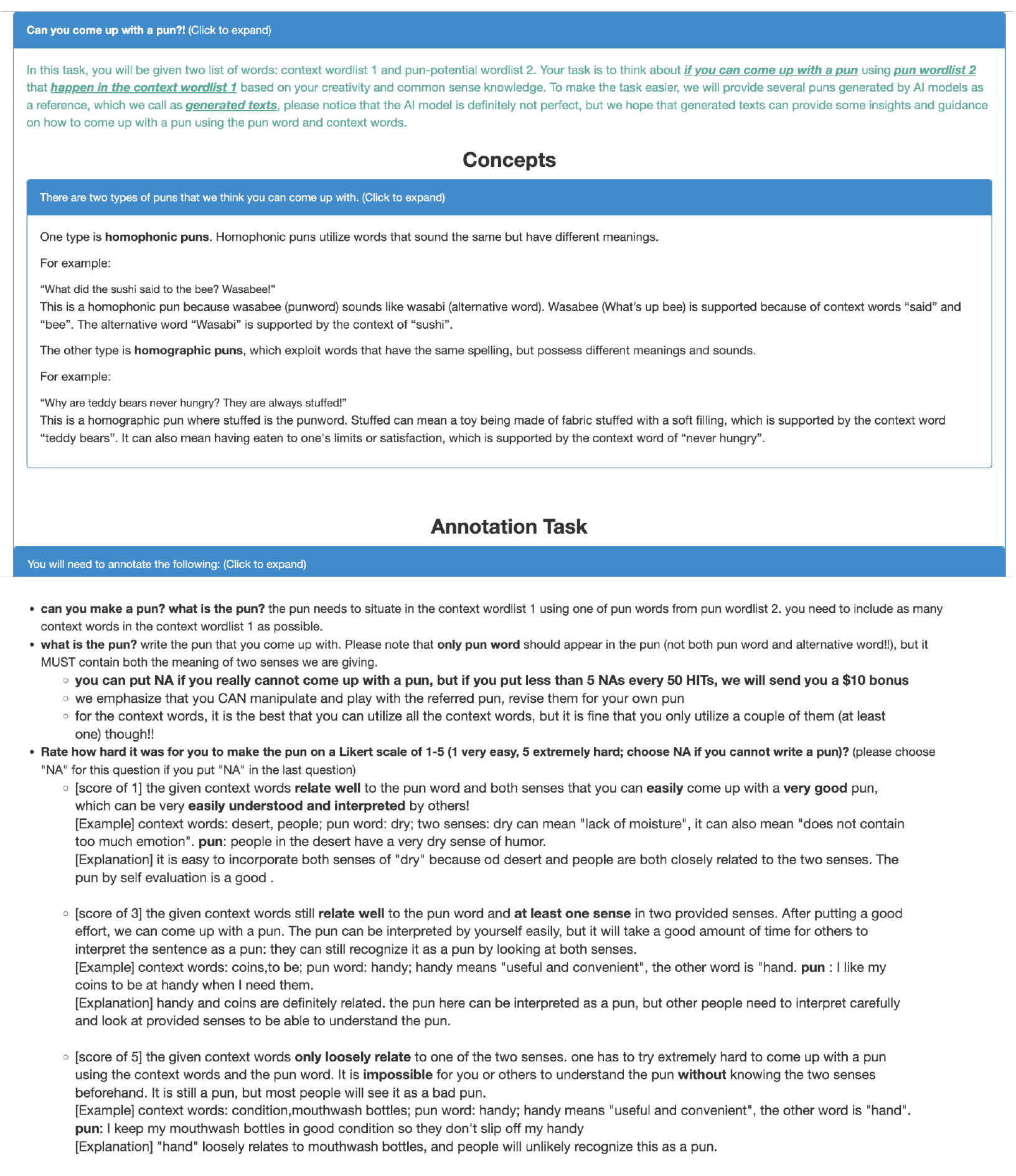}
    % \vspace{-0.25in}
    \caption{The annotation interface for collecting CUP dataset.}
    \label{fig:annotation-guideline}
    % \vspace{-0.2in}
\end{figure*}

Figure~\ref{fig:annotation-guideline} shows our annotation interface for collecting the CUP dataset.